\documentclass{ifacconf}
\usepackage{graphicx}      
\usepackage{natbib}
\usepackage{amsmath, amssymb, bm}
\let\theoremstyle\relax      
\usepackage{amsthm}
\usepackage{mathtools} 
\usepackage{physics}   
\usepackage{natbib}    
\usepackage{algorithm}
           
\usepackage{algorithmic}
\usepackage{tikz}
\usetikzlibrary{arrows.meta, shapes, positioning} 
\usepackage{booktabs}
\usepackage{url}

\newtheorem{definition}{Definition}
\newcommand{\khatri}{\odot}          
\newcommand{\hadamard}{\circledast}  
\newtheorem*{remark}{Remark}
\begin{document}
\begin{frontmatter}

\title{A Fully Probabilistic Tensor Network for Regularized Volterra System Identification} 

\thanks[footnoteinfo]{This publication is part of the project \textit{Sustainable Learning for AI from Noisy Large-Scale Data} (project number VI.Vidi.213.017), which is financed by the Dutch Research Council (NWO).}
\vspace{-0.3cm}
\author[First]{Afra Kilic} 
\author[First]{Kim Batselier} 

\vspace{-0.3cm}
\address[First]{Delft University of Technology , Delft, Netherlands
\newline (e-mail: \{h.a.kilic, k.batselier\}@ tudelft.nl).}

\vspace{-0.3cm}

\begin{abstract}
Modeling nonlinear systems with Volterra series is challenging since the number of kernel coefficients grows exponentially with the model order. This work introduces Bayesian Tensor Network Volterra kernel machines (BTN-V), extending the Bayesian Tensor Network (BTN) framework to Volterra system identification. BTN-V represents Volterra kernels via canonical polyadic decomposition, reducing model complexity from $\mathcal{O}(I^D)$ to $\mathcal{O}(DIR)$. By treating all tensor components and hyperparameters as random variables, BTN-V provides predictive uncertainty estimation at no extra computational cost. Sparsity-inducing hierarchical priors enable automatic rank determination and learning of fading-memory behavior directly from data, improving interpretability and avoiding overfitting. Empirical results demonstrate competitive accuracy, enhanced uncertainty quantification, and reduced computational cost.
\end{abstract}

\begin{keyword}
nonlinear system identification \sep Volterra series \sep tensor network kernel machines \sep variational inference \sep Bayesian methods
\end{keyword}

\end{frontmatter}

\section{Introduction}

Modeling nonlinear systems remains a key challenge in system identification. The Volterra series, a nonlinear extension of the finite impulse response model, is commonly used when the exact system behavior is unknown. However, its application is typically limited to weakly nonlinear systems, where linear dynamics dominate. This limitation arises not from the Volterra framework itself but from the exponential growth of kernel coefficients with increasing model order.

Several methods exist to reduce the complexity of Volterra kernels. One approach incorporates prior knowledge of the kernel structure within a Bayesian framework \citep{chen_estimation_2011}, encoding the expected smooth decay of Volterra kernels via Bayesian priors \citep{pillonetto_new_2010}. This was extended to parametric Volterra models with decaying covariance matrices \citep{birpoutsoukis_efficient_2018}, though limited to third-order kernels, while higher-order non-parametric models using similar structures remain restricted to small datasets \citep{DallaLibera2021}.

An alternative approach imposes sparsity by assuming most Volterra coefficients are negligible, implemented through sparse Bayesian learning \citep{miao_kernels_2019}. However, this strong sparsity assumption applies only to the specific system considered in that study. A further alternative is kernel compression using low-rank tensor structures, which represent all Volterra coefficients through a small set of parameters from which the full set can be reconstructed. \cite{favier2012nonlinear} apply both low-rank Tucker and canonical polyadic decompositions. \cite{batselier_tensor_2017} has proposed a multiple-input–multiple-output (MIMO) Volterra tensor network (TN) that uses tensor-train decomposition to represent all Volterra kernels at once. By exploiting multilinearity and applying alternating least squares (ALS), the MIMO Volterra TN avoids the exponential growth of coefficients and enables efficient identification of high-order (up to 10th-order) Volterra systems within seconds on standard hardware.

Although the MIMO Volterra TN offers low computational complexity, it does not account for parameter uncertainty and therefore cannot provide uncertainty quantification for predictions. It also requires the tensor rank to be specified a priori and tends to overfit when higher ranks are used. Rank selection is often performed via trial and error, which is both costly and imprecise, while more systematic approaches, such as maximum likelihood estimation, may still lead to overfitting, and cross-validation becomes computationally expensive when multiple hyperparameters are involved. The first probabilistic treatment of the Volterra TN with Tikhonov regularization has been proposed to avoid overfitting and to quantify uncertainty in the predictions \citep{memmel_bayesian_2023}; however, it models only a single TN core as a random variable, treating the remaining cores as deterministic, and is thus not fully probabilistic. Moreover, despite the regularization, this approach still requires manual tuning of the tensor rank. Recently, \citet{kilic_interpretable_2025} presented the Bayesian Tensor Network (BTN) kernel machines framework, a fully probabilistic approach that employs sparsity-inducing hierarchical priors on the TN components to automatically infer model complexity. Specifically, these sparsity-inducing priors allow the model, during identification, to automatically determine the effective tensor rank and identify the most relevant features for prediction, thereby enhancing interpretability. 

\noindent \textbf{In this paper}, we propose the Bayesian Tensor Network Volterra kernel machines (BTN-V) by extending the BTN kernel machines framework \citep{kilic_interpretable_2025} to Volterra system identification. BTN-V treats all tensor components and model hyperparameters as random variables, allowing prediction uncertainty to be estimated without additional computational cost. Through the use of sparsity-inducing priors, BTN-V offers two main benefits: it automatically determines the tensor rank during the identification process and learns the fading-memory behavior of Volterra systems directly from the data, helping to prevent overfitting by reducing the influence of distant past inputs.

\section{Tensor Basics and Notation}

The order of a tensor is the number of its dimensions, also reffered to as ways or modes. Scalars are denoted by lowercase letters, e.g., $a$; vectors (first-order tensors) by bold lowercase letters, e.g., $\mathbf{a}$; matrices (second-order tensors) by bold uppercase letters, e.g., $\mathbf{A}$; and higher-order tensors (order $\geq 3$) by bold calligraphic letters, e.g., $\bm{\mathcal{A}}$.  

A $D$th-order tensor $\bm{\mathcal{A}} \in \mathbb{R}^{I_1 \times I_2 \times \cdots \times I_D}$ has entries $a_{i_1 i_2 \cdots i_D}$, with indices $i_d = 1, \dots, I_d$ for $d \in [1,D]$. Often, it is computationally convenient to vectorize tensors. The vectorization $\operatorname{vec}(\bm{\mathcal{A}}) \in \mathbb{R}^{I_1 I_2 \cdots I_D}$ maps each entry as  \vspace{-0.5cm}

{\fontsize{9}{11}\selectfont
\begin{equation}
\operatorname{vec}(\bm{\mathcal{A}})_i = a_{i_1 i_2 \dots i_D}, \quad
i = i_1 + \sum_{d=2}^{D} (i_d - 1) \prod_{k=1}^{d-1} I_k.
\end{equation}
}

When applied to a matrix, the operator $\operatorname{vec}(\cdot)$ performs column-wise vectorization.  The operator $\operatorname{diag}(\cdot)$ returns a diagonal matrix from a vector. The Kronecker product of $\mathbf{A} \in \mathbb{R}^{I \times J}$ and $\mathbf{B} \in \mathbb{R}^{K \times L}$ is $\mathbf{A} \otimes \mathbf{B} \in \mathbb{R}^{KI \times LJ}$. The Khatri-Rao product $\mathbf{A} \khatri \mathbf{B}$ of $\mathbf{A} \in \mathbb{R}^{I \times J}$ and $\mathbf{B} \in \mathbb{R}^{K \times J}$ is an $IK \times J$ matrix formed by column-wise Kronecker products. The Hadamard (element-wise) product of $\mathbf{A}, \mathbf{B} \in \mathbb{R}^{I \times J}$ is $\mathbf{A} \hadamard \mathbf{B} \in \mathbb{R}^{I \times J}$. The identity matrix is denoted by $\mathbf{I}$, with size either inferred or specified.

\vspace{1mm}
\begin{definition}
A rank-$R$ Canonical Polyadic Decomposition (CPD) of $\mathbf{w} = \operatorname{vec}(\bm{\mathcal{W}}) \in \mathbb{R}^{I^D}$ consists of $D$ factor matrices $\mathbf{W}^{(d)} \in \mathbb{R}^{I \times R}$, such that

\vspace{-0.5cm} {\fontsize{9}{11}\selectfont \begin{equation} \mathbf{w} =(\mathbf{W}^{(1)} \odot \mathbf{W}^{(2)} \odot \dots \odot \mathbf{W}^{(D)}) \bm{1}_{R} = \sum_{r=1}^{R} \mathbf{w}^{(1)}_r \otimes \mathbf{w}^{(2)}_r \otimes \dots \otimes \mathbf{w}^{(D)}_r, \label{eq1} \end{equation}} \vspace{-0.2cm}

\label{def:1}
where $\mathbf{w}^{(d)}_r \in \mathbb{R}^{I}$ denotes the $r$th column of the matrix $\mathbf{W}^{(d)}$. The CPD is unique under mild conditions \citep{kruskal1977three}, and its storage complexity scales as $\mathcal{O}(DIR)$.
\end{definition}

\section{Tensor Network SISO Volterra System}

In this section, we briefly introduce the TN formulation of the single-input single-output (SISO) Volterra system \citep{batselier_tensor_2017}. We focus on the SISO case for simplicity; the extension to MIMO is trivial but not discussed for brevity.

Consider a discrete time truncated $D$th order Volterra system with memory length $M$:

\vspace{-0.5cm}

{\fontsize{9}{11}\selectfont
\begin{equation}
y(n) = \sum_{d=0}^{D} \sum_{m_1, \ldots, m_d = 0}^{M-1} \bm{\mathcal{W}}_d(m_1, \ldots, m_d) \prod_{j=1}^{d} u(n - m_j) + e(n), 
\end{equation}}

where $u(n)$ is the input, $y(n)$ is the output, and $e(n)$ represents additive noise, for for $n = 1, \ldots, N$. The set $\{\bm{\mathcal{W}}_d\}_{d=0}^D$ represents the Volterra kernels of different orders. These kernels grow exponentially with the system order $D$. This curse of dimensionality can be mitigated by representing all Volterra kernels simultaneously with a low-rank TN \citep{batselier_tensor_2017}, which transforms the problem into a set of smaller linear systems, each solved iteratively during the identification process. To obtain a low-rank TN representation of the Volterra kernels, the first step is to rewrite the inputs $u(n)$ following \cite{batselier_tensor_2017} as

\vspace{-0.5cm}
{\fontsize{9}{11}\selectfont
\begin{equation}
    \mathbf{u}_n =
    \begin{pmatrix}
        1 & u(n) & \cdots & u(n - M + 1)
    \end{pmatrix}^\top \in \mathbb{R}^I,
    \label{eq2}
\end{equation}}

\vspace{-0.25cm}

where $I := M + 1$, with the corresponding system output is $y(n) \in \mathbb{R}$. Including the constant term as the first term in $\bm{u}_n$ allows us to define a feature map $\mathbf{u}_n^{D}$ as the $D$-times Kronecker product of $\mathbf{u}_n$, such that 

\vspace{-0.5cm}

{\fontsize{9}{11}\selectfont
\begin{equation}
    \mathbf{u}_n^{D} :=
    \underbrace{\mathbf{u}_n \otimes \mathbf{u}_n \otimes \cdots \otimes \mathbf{u}_n}_{D \text{ times}}
    \in \mathbb{R}^{I^D},
    \label{eq3}
\end{equation}}

\vspace{-0.25cm}

which contains all monomials of the input from degree $0$ up to $D$. With this definition, the output at time $n$ can be expressed as

\vspace{-0.5cm}
{\fontsize{9}{11}\selectfont
\begin{equation}
    {y}(n) = \left( \mathbf{u}_n^{D} \right)^\top \mathbf{w} + e_n,
    \label{eq4}
\end{equation}}

\vspace{-0.3cm}

where $\mathbf{w} \in \mathbb{R}^{I^D}$ vector containing all coefficients from all Volterra kernelss. Since the Volterra feature map $\mathbf{u}_n^{D}$ in \eqref{eq3} has a Kronecker structure, this property can be exploited to express the Volterra coefficients $\mathbf{w}$ using the CPD model in Definition~\ref{def:1}, which reduces the number of learnable parameters from $\mathcal{O}(I^D)$ to $\mathcal{O}(DIR)$ \citep{harshman1970foundations}. In this model, $\mathbf{w}$ is represented by $D$ factor matrices of size $\mathbb{R}^{I \times R}$, which are estimated iteratively using the ALS algorithm without explicitly constructing $\mathbf{w}$. When the rank $R$ matches the true CP rank, the exact solution of \eqref{eq4} is recovered.  However, ALS assumes deterministic factor matrices and yields only point estimates. Following \cite{kilic_interpretable_2025}, we instead treat the factor matrices as random variables with sparsity-inducing priors. This probabilistic formulation allows uncertainty quantification in the predictions and the model to automatically infer both the effective tensor rank $R$ and the fading-memory behavior of the Volterra series during identification.
 
\section{Probabilistic Tensor Network SISO Volterra Systems}

In this section, we present the main contribution of this paper: BTN-V, the BTN kernel machines framework extended to SISO Volterra systems. For a detailed discussion of the priors and the mean-field variational inference-based probabilistic identification procedure, the reader is referred to \citet{kilic_interpretable_2025}.

\subsection{Probabilistic Model and Priors}
Expressing equation~\eqref{eq4} for $n = 0, 1, \ldots, N$ leads to the following matrix equation:

\vspace{-0.5cm}
{\fontsize{9}{11}\selectfont
\begin{equation}
    \bm{y} = \bm{U}^{D^T} \bm{w} + \bm{e},
    \label{eq5}
\end{equation}}

where 
$\bm{y} \in \mathbb{R}^{N}$ is the vector of measured outputs and $\bm{e} \in \mathbb{R}^{N}$ denotes Gaussian white noise with $\bm{e} \sim \mathcal{N}(\bm{0}, \tau^{-1} \mathbf{I}_N)$. Let $\mathbf{U} \in \mathbb{R}^{I \times N}$, where the $n$th column contains $\mathbf{u}_n$. The matrix 
$\mathbf{U}^{D} =\mathbf{U} \,\khatri\, \mathbf{U} \,\khatri\, \cdots \,\khatri\, \mathbf{U} \in \mathbb{R}^{I^{D} \times N}$ 
represents the $D$-fold row-wise Khatri–Rao product of $\mathbf{U}$ with itself, thus $\bm{U}^D$ the input matrix whose $n$th column corresponds to the Kronecker vector $\bm{u}_n^{D}$. Assuming a CP-decomposed-Volterra-coefficients vector $\mathbf{w}$ together with the Gaussian noise model, the likelihood of the observed outputs is given by

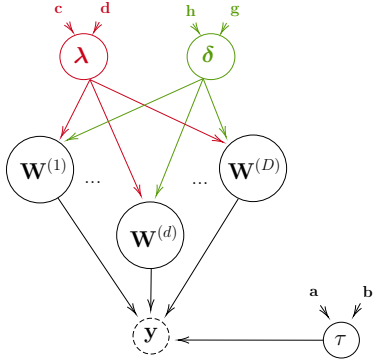
\begin{figure}[t]
\centering 
\scalebox{0.50}{
\begin{tikzpicture}[x=0.75pt,y=0.75pt,yscale=-1,xscale=1]
\draw   (188.67,173.57) .. controls (188.67,155.07) and (203.67,140.07) .. (222.17,140.07) .. controls (240.67,140.07) and (255.67,155.07) .. (255.67,173.57) .. controls (255.67,192.08) and (240.67,207.07) .. (222.17,207.07) .. controls (203.67,207.07) and (188.67,192.08) .. (188.67,173.57) -- cycle ;
\draw   (298.67,239.57) .. controls (298.67,221.07) and (313.67,206.07) .. (332.17,206.07) .. controls (350.67,206.07) and (365.67,221.07) .. (365.67,239.57) .. controls (365.67,258.08) and (350.67,273.07) .. (332.17,273.07) .. controls (313.67,273.07) and (298.67,258.08) .. (298.67,239.57) -- cycle ;
\draw   (401.33,172.24) .. controls (401.33,153.74) and (416.33,138.74) .. (434.83,138.74) .. controls (453.33,138.74) and (468.33,153.74) .. (468.33,172.24) .. controls (468.33,190.74) and (453.33,205.74) .. (434.83,205.74) .. controls (416.33,205.74) and (401.33,190.74) .. (401.33,172.24) -- cycle ;
\draw [dash pattern={on 4pt off 2pt}]  (314.99,341.18) .. controls (314.94,331.23) and (322.96,323.13) .. (332.9,323.09) .. controls (342.84,323.04) and (350.94,331.06) .. (350.99,341) .. controls (351.04,350.94) and (343.02,359.04) .. (333.08,359.09) .. controls (323.14,359.14) and (315.04,351.12) .. (314.99,341.18) -- cycle ;
\draw   (504.99,345.18) .. controls (504.94,335.23) and (512.96,327.13) .. (522.9,327.09) .. controls (532.84,327.04) and (540.94,335.06) .. (540.99,345) .. controls (541.04,354.94) and (533.02,363.04) .. (523.08,363.09) .. controls (513.14,363.14) and (505.04,355.12) .. (504.99,345.18) -- cycle ;
\draw    (240,202) -- (316.89,317.34) ;
\draw [shift={(318,319)}, rotate = 236.31] [color={rgb, 255:red, 0; green, 0; blue, 0 }  ][line width=0.75]    (10.93,-3.29) .. controls (6.95,-1.4) and (3.31,-0.3) .. (0,0) .. controls (3.31,0.3) and (6.95,1.4) .. (10.93,3.29)   ;
\draw    (420,203) -- (349.05,318.3) ;
\draw [shift={(348,320)}, rotate = 301.61] [color={rgb, 255:red, 0; green, 0; blue, 0 }  ][line width=0.75]    (10.93,-3.29) .. controls (6.95,-1.4) and (3.31,-0.3) .. (0,0) .. controls (3.31,0.3) and (6.95,1.4) .. (10.93,3.29)   ;
\draw    (504.99,345.18) -- (362,345) ;
\draw [shift={(360,345)}, rotate = 0.07] [color={rgb, 255:red, 0; green, 0; blue, 0 }  ][line width=0.75]    (10.93,-3.29) .. controls (6.95,-1.4) and (3.31,-0.3) .. (0,0) .. controls (3.31,0.3) and (6.95,1.4) .. (10.93,3.29)   ;
\draw    (333.17,273.07) -- (332.69,313.07) ;
\draw [shift={(332.67,315.07)}, rotate = 270.68] [color={rgb, 255:red, 0; green, 0; blue, 0 }  ][line width=0.75]    (10.93,-3.29) .. controls (6.95,-1.4) and (3.31,-0.3) .. (0,0) .. controls (3.31,0.3) and (6.95,1.4) .. (10.93,3.29)   ;
\draw    (502,310.07) -- (508.89,320.41) ;
\draw [shift={(510,322.07)}, rotate = 236.31] [color={rgb, 255:red, 0; green, 0; blue, 0 }  ][line width=0.75]    (10.93,-3.29) .. controls (6.95,-1.4) and (3.31,-0.3) .. (0,0) .. controls (3.31,0.3) and (6.95,1.4) .. (10.93,3.29)   ;
\draw    (542,308.07) -- (535.95,319.31) ;
\draw [shift={(535,321.07)}, rotate = 298.3] [color={rgb, 255:red, 0; green, 0; blue, 0 }  ][line width=0.75]    (10.93,-3.29) .. controls (6.95,-1.4) and (3.31,-0.3) .. (0,0) .. controls (3.31,0.3) and (6.95,1.4) .. (10.93,3.29)   ;
\draw [color={rgb, 255:red, 87; green, 155; blue, 9 }  ,draw opacity=1 ]   (374.64,26.73) -- (380.07,34.81) ;
\draw [shift={(381.18,36.47)}, rotate = 236.08] [color={rgb, 255:red, 87; green, 155; blue, 9 }  ,draw opacity=1 ][line width=0.75]    (10.93,-3.29) .. controls (6.95,-1.4) and (3.31,-0.3) .. (0,0) .. controls (3.31,0.3) and (6.95,1.4) .. (10.93,3.29)   ;
\draw [color={rgb, 255:red, 87; green, 155; blue, 9 }  ,draw opacity=1 ]   (408.36,26.11) -- (403.59,34.9) ;
\draw [shift={(402.64,36.66)}, rotate = 298.5] [color={rgb, 255:red, 87; green, 155; blue, 9 }  ,draw opacity=1 ][line width=0.75]    (10.93,-3.29) .. controls (6.95,-1.4) and (3.31,-0.3) .. (0,0) .. controls (3.31,0.3) and (6.95,1.4) .. (10.93,3.29)   ;
\draw  [color={rgb, 255:red, 87; green, 155; blue, 9 }  ,draw opacity=1 ] (369,59.91) .. controls (369,47.16) and (379.67,36.82) .. (392.83,36.82) .. controls (406,36.82) and (416.67,47.16) .. (416.67,59.91) .. controls (416.67,72.66) and (406,83) .. (392.83,83) .. controls (379.67,83) and (369,72.66) .. (369,59.91) -- cycle ;
\draw [color={rgb, 255:red, 202; green, 4; blue, 28 }  ,draw opacity=1 ]   (248.64,25.73) -- (254.07,33.81) ;
\draw [shift={(255.18,35.47)}, rotate = 236.08] [color={rgb, 255:red, 202; green, 4; blue, 28 }  ,draw opacity=1 ][line width=0.75]    (10.93,-3.29) .. controls (6.95,-1.4) and (3.31,-0.3) .. (0,0) .. controls (3.31,0.3) and (6.95,1.4) .. (10.93,3.29)   ;
\draw [color={rgb, 255:red, 202; green, 4; blue, 28 }  ,draw opacity=1 ]   (281.36,25.11) -- (276.59,33.9) ;
\draw [shift={(275.64,35.66)}, rotate = 298.5] [color={rgb, 255:red, 202; green, 4; blue, 28 }  ,draw opacity=1 ][line width=0.75]    (10.93,-3.29) .. controls (6.95,-1.4) and (3.31,-0.3) .. (0,0) .. controls (3.31,0.3) and (6.95,1.4) .. (10.93,3.29)   ;
\draw  [color={rgb, 255:red, 202; green, 4; blue, 28 }  ,draw opacity=1 ] (242,59.91) .. controls (242,47.16) and (252.67,36.82) .. (265.83,36.82) .. controls (279,36.82) and (289.67,47.16) .. (289.67,59.91) .. controls (289.67,72.66) and (279,83) .. (265.83,83) .. controls (252.67,83) and (242,72.66) .. (242,59.91) -- cycle ;
\draw [color={rgb, 255:red, 202; green, 4; blue, 28 }  ,draw opacity=1 ]   (272,83) -- (322.2,198.17) ;
\draw [shift={(323,200)}, rotate = 246.45] [color={rgb, 255:red, 202; green, 4; blue, 28 }  ,draw opacity=1 ][line width=0.75]    (10.93,-3.29) .. controls (6.95,-1.4) and (3.31,-0.3) .. (0,0) .. controls (3.31,0.3) and (6.95,1.4) .. (10.93,3.29)   ;
\draw [color={rgb, 255:red, 87; green, 155; blue, 9 }  ,draw opacity=1 ]   (385,82) -- (338.74,198.14) ;
\draw [shift={(338,200)}, rotate = 291.72] [color={rgb, 255:red, 87; green, 155; blue, 9 }  ,draw opacity=1 ][line width=0.75]    (10.93,-3.29) .. controls (6.95,-1.4) and (3.31,-0.3) .. (0,0) .. controls (3.31,0.3) and (6.95,1.4) .. (10.93,3.29)   ;
\draw [color={rgb, 255:red, 202; green, 4; blue, 28 }  ,draw opacity=1 ]   (272,83) -- (403.19,145.14) ;
\draw [shift={(405,146)}, rotate = 205.35] [color={rgb, 255:red, 202; green, 4; blue, 28 }  ,draw opacity=1 ][line width=0.75]    (10.93,-3.29) .. controls (6.95,-1.4) and (3.31,-0.3) .. (0,0) .. controls (3.31,0.3) and (6.95,1.4) .. (10.93,3.29)   ;
\draw [color={rgb, 255:red, 87; green, 155; blue, 9 }  ,draw opacity=1 ]   (385,82) -- (250.82,143.17) ;
\draw [shift={(249,144)}, rotate = 335.49] [color={rgb, 255:red, 87; green, 155; blue, 9 }  ,draw opacity=1 ][line width=0.75]    (10.93,-3.29) .. controls (6.95,-1.4) and (3.31,-0.3) .. (0,0) .. controls (3.31,0.3) and (6.95,1.4) .. (10.93,3.29)   ;
\draw [color={rgb, 255:red, 87; green, 155; blue, 9 }  ,draw opacity=1 ]   (385,82) -- (407.27,139.14) ;
\draw [shift={(408,141)}, rotate = 248.7] [color={rgb, 255:red, 87; green, 155; blue, 9 }  ,draw opacity=1 ][line width=0.75]    (10.93,-3.29) .. controls (6.95,-1.4) and (3.31,-0.3) .. (0,0) .. controls (3.31,0.3) and (6.95,1.4) .. (10.93,3.29)   ;
\draw [color={rgb, 255:red, 202; green, 4; blue, 28 }  ,draw opacity=1 ]   (272,83) -- (243.89,139.21) ;
\draw [shift={(243,141)}, rotate = 296.57] [color={rgb, 255:red, 202; green, 4; blue, 28 }  ,draw opacity=1 ][line width=0.75]    (10.93,-3.29) .. controls (6.95,-1.4) and (3.31,-0.3) .. (0,0) .. controls (3.31,0.3) and (6.95,1.4) .. (10.93,3.29)   ;

\draw (515,338.47) node [anchor=north west][inner sep=0.75pt]    {\Large$\tau $};
\draw (325,331.47) node [anchor=north west][inner sep=0.75pt]    {\Large$\mathbf{y}$};
\draw (310.67,228.48) node [anchor=north west][inner sep=0.75pt]  [rotate=-359.99]  {\Large${\displaystyle \mathbf{W}^{(d)}}$};
\draw (200.67,160.48) node [anchor=north west][inner sep=0.75pt]  [rotate=-359.99]  {\Large${\displaystyle \mathbf{W}^{(1)}}$};
\draw (408.33,160.64) node [anchor=north west][inner sep=0.75pt]  [rotate=-359.99]  {\Large ${\displaystyle \mathbf{W}^{(D)}}$};
\draw (490,292.07) node [anchor=north west][inner sep=0.75pt]   [align=left] {\large $\mathbf{a}$};
\draw (543,289.07) node [anchor=north west][inner sep=0.75pt]   [align=left] {\large $\mathbf{b}$};
\draw (366.32,7.09) node [anchor=north west][inner sep=0.75pt]  [color={rgb, 255:red, 87; green, 155; blue, 9 }  ,opacity=1 ] [align=left] {\large $\mathbf{h}$};
\draw (411.18,8.09) node [anchor=north west][inner sep=0.75pt]  [color={rgb, 255:red, 87; green, 155; blue, 9 }  ,opacity=1 ] [align=left] {\large $\mathbf{g}$};
\draw (382.27,48.34) node [anchor=north west][inner sep=0.75pt]  [color={rgb, 255:red, 87; green, 155; blue, 9 }  ,opacity=1 ]  {\Large$\bm{\delta}$};
\draw (235.32,8.09) node [anchor=north west][inner sep=0.75pt]  [color={rgb, 255:red, 202; green, 4; blue, 28 }  ,opacity=1 ] [align=left] {\large $\mathbf{c}$};
\draw (281.18,4.09) node [anchor=north west][inner sep=0.75pt]  [color={rgb, 255:red, 202; green, 4; blue, 28 }  ,opacity=1 ] [align=left] {\large $\mathbf{d}$};
\draw (254.27,49.34) node [anchor=north west][inner sep=0.75pt]  [color={rgb, 255:red, 202; green, 4; blue, 28 }  ,opacity=1 ]  {\Large$\bm{\lambda}$};
\draw (265,184) node [anchor=north west][inner sep=0.75pt]   [align=left] {{\LARGE ...}};
\draw (372,185) node [anchor=north west][inner sep=0.75pt]   [align=left] {{\LARGE ...}};

\end{tikzpicture}}
\caption{\footnotesize
Probabilistic graphical model of the Volterra TN, where the CPD-decomposed coefficients $\mathbf{w}$ are represented by factor matrices $\{\mathbf{W}^{(d)}\}_{d=1}^D$. Dashed, solid, and unbounded nodes denote observed data $\mathbf{y}$, random variables, and Gamma hyperparameters, respectively..
}

    \label{fig:CPD_model}

\end{figure}

\vspace{-0.5cm}

{\fontsize{9}{11}\selectfont
\begin{equation}
    p \left(\mathbf{y} \mid \{\mathbf{W}^{(d)}\}_{d=1}^{D}, \tau \right) = \prod_{n=1}^{N} \mathcal{N}\left(y(n) \mid \left( \bm{u}_n^{D} \right)^\top\mathbf{w}, \tau^{-1} \right),
    \label{eq6}
\end{equation}}

\vspace{-0.2cm}

where $\tau$ denotes the noise precision and $\mathbf{W}^{(d)} \in \mathbb{R}^{I\times R}$ are the factor matrices of the CP-decomposed model weights (Definition~\ref{def:1}), whose $r$th column is $\mathbf{w}_r^{(d)}$ and $i$th row is $\mathbf{w}_i^{(d)}$, with $r = 1,\dots,R$ and $i = 1,\dots,I$. Determining an appropriate tensor rank $R$ and feature dimension $I$ is generally a nontrivial and computationally intensive task. 

To address this, \citet{kilic_interpretable_2025} use a hierarchical sparsity-inducing prior to automatically infer $R$ and $I$, avoiding overfitting. Building on this methodology, we define two sets of sparsity parameters: $\bm{\lambda} := [\lambda_1, \dots, \lambda_R]$, with each $\lambda_r$ controlling the regularization of the $r$th \emph{column} $\mathbf{w}_r^{(d)}$, and $\bm{\delta} := [\delta_{1}, \dots, \delta_{I}]$, with each $\delta_i$ regulating the $i$th \emph{row} $\mathbf{w}_i^{(d)}$ of $\mathbf{W}^{(d)}$, for all $d \in [1, D]$. Together, $\bm{\lambda}$ and $\bm{\delta}$ govern sparsity across the columns and rows of the factor matrices, respectively. The prior distribution for the vectorized factor matrices is a zero mean Gaussian prior

\vspace{-0.5cm}
{\fontsize{9}{11}\selectfont
\begin{equation}
    p\left(\operatorname{vec}(\mathbf{W}^{(d)}) \mid \bm{\lambda}, \bm{\delta}\right)
= \mathcal{N}\!\left(\mathbf{0},\, \bm{\Lambda}^{-1} \otimes \bm{\Delta}^{-1}\right)
\label{eq7}
\end{equation}
}

where $\bm{\Lambda} \otimes \bm{\Delta} = \operatorname{diag}(\bm{\lambda}) \otimes \operatorname{diag}(\bm{\delta})$ represents the inverse covariance matrix (precision matrix). Under the zero-mean prior assumption, higher values in $\bm{\lambda}$ or $\bm{\delta}$ force the corresponding columns or rows of $\mathbf{W}^{(d)}$ toward zero, thereby regularizing the associated components or input dimensions. This follows the automatic relevance determination (ARD) principle \citep{neal1996bayesian} and its tensor-based extensions \citep{Zhao_2015_rank_det}. 

For a fully probabilistic formulation, we further specify Gamma hyperpriors over the sparsity parameters as $p(\bm{\lambda}) = \prod_{r=1}^{R} \text{Ga}(\lambda_r \mid c_0, d_0)$ and $p(\bm{\delta}) = \prod_{i=1}^{I} \text{Ga}(\delta_i \mid g_0, h_0)$, where $\text{Ga}(x \mid a, b)$ denotes the Gamma distribution. Similarly, the noise precision $\tau$ is assigned a Gamma prior $p(\tau) = \text{Ga}(\tau \mid a_0, b_0)$. 

\begin{remark} In the original BTN-Kernel machines formulation, a separate $\bm{\delta}_{d}$ was defined for each factor matrix $\mathbf{W}^{(d)}$, since the feature map was formed by the Kronecker product of $D$ different feature vectors $\mathbf{u}^{(d)}_n$. In contrast, the Volterra feature map in \eqref{eq3} uses the same input vector $\mathbf{u}_n$ repeated $D$ times in the Kronecker product. Thus, we use a single vector $\bm{\delta}$ that applies row-wise regularization across all factor matrices. Since each row corresponds to a specific memory lag, $\bm{\delta}$ is directly associated with the temporal structure of the model (i.e., the memory lags). By modeling $\bm{\delta}$ as a random variable, the model learns which lags are more important during identification, thereby regularizing the temporal structure by reducing the influence of less informative lags and highlighting those that are more relevant. 
\end{remark}

Denoting all latent variables and hyperparameters by $\Theta = \{\mathbf{W}^{(1)}, \ldots, \mathbf{W}^{(D)}, \bm{\delta}, \bm{\lambda}, \tau\}$, the joint distribution $p(\mathbf{y}, \Theta)$ can be expressed as

\vspace{-0.5cm}

{\fontsize{9}{11}\selectfont
\begin{equation}
p \left( \mathbf{y} \mid \{\mathbf{W}^{(d)}\}_{d=1}^{D}, \tau \right)
\prod_{d=1}^{D} p\left(\mathbf{W}^{(d)} \mid \bm{\lambda}, \bm{\delta}\right)\, p(\bm{\delta})\, p(\bm{\lambda})\, p(\tau).
\label{eq8}
\end{equation}
}

Figure \ref{fig:CPD_model} illustrates the probabilistic graphical model corresponding to the joint distribution in \eqref{eq8}. The input matrices $\bm{U}$ are omitted from the figure since they are not random variables. The factor matrices $\{\mathbf{W}^{(d)}\}_{d=1}^D$, and the shape and scale hyperparameters of the Gamma distributions, represented by unbounded nodes in the figure, must be initialized before identification. Then the full posterior distribution of all variables in \( \Theta \) given the observed data is 
{\fontsize{9}{11}\selectfont
\begin{equation}
    p(\Theta | \mathbf{y}) = \frac{p(\mathbf{y}, \Theta)}{\int p(\mathbf{y}, \Theta) d \Theta}. 
    \label{eq:9}
\end{equation}}
Based on the posterior distribution of $\Theta$, the predictive distribution over unseen data points, denoted $\tilde{y_i}$ can be inferred by 

\vspace{-0.5cm}

{\fontsize{9}{11}\selectfont
\begin{equation}
     p(\tilde{y_i} \mid \mathbf{y}) = \int p\left(\tilde{y_i} \mid \Theta\right) p\left(\Theta \mid \mathbf{y} \right)d\Theta.
     \label{eq:10} 
\end{equation}}

\vspace{-0.2cm}

\subsection{Identification Process}

An exact Bayesian inference of \eqref{eq:9} and \eqref{eq:10} requires integrating over all latent variables and hyperparameters, making it analytically intractable. Therefore, as presented in \citep{kilic_interpretable_2025} we perform a Bayesian identification using mean-field variational inference \citep{winn_variational_2005}. The key idea is to approximate the true posterior $p(\Theta \mid \mathbf{y})$ with a variational distribution $q(\Theta)$ by minimizing the Kullback--Leibler (KL) divergence. This is equivalent to maximizing a lower bound $\mathcal{L}(q)$ on the model evidence $\ln p(\mathbf{y})$, defined as $\mathcal{L}(q) = \mathbb{E}_{q(\Theta)}[\ln p(\mathbf{y},\Theta)]$. The maximum of the lower bound occurs when the KL divergence vanishes, implying \( q(\Theta) = p(\Theta \mid \mathbf{y}) \). The mean-field assumption factorizes the variational distribution over each variable $\theta_j \in \bm{\Theta}$ as

\vspace{-0.5cm}

{\fontsize{9}{11}\selectfont
\begin{equation}
    q(\Theta) = \prod_{d=1}^{D} q_{\mathbf{W}^{(d)}}(\mathbf{W}^{(d)}) \, q_{\bm{\delta}}(\bm{\delta}) \, q_{\bm{\lambda}}(\bm{\lambda}) \, q_{\tau}(\tau).
    \label{eq11}
\end{equation}}

In other words, this factorization assumes that all variables $\theta_j \in \bm{\Theta}$ are independent. The functional form of each factor can then be derived analytically in turn. Specifically, the optimal $\theta_j$ is obtained by maximizing $\mathcal{L}(q)$, with the maximum occurring when
{\fontsize{9}{11}\selectfont
\begin{equation}
    \ln q_j(\theta_j) = \mathbb{E}_{q(\Theta \setminus \theta_j)}[\ln p(\mathbf{y}, \Theta)] + \text{const},
    \label{eq12}
\end{equation}}

\vspace{-0.25cm}

where \( \mathbb{E}_{q(\Theta \setminus \theta_j)} [\cdot] \) denotes the expectation taken with respect to the variational distributions of all variables except \( \theta_j \). Since all parameter distributions are in the exponential family and conjugate to their priors, this yields closed-form posterior updates for each factor. Identification proceeds by initializing each $q_j(\theta_j)$ and iteratively updating them using the update rules until convergence. In the following, we present the posterior update rules for each $q_j(\theta_j)$, for all $\theta_j \in \bm{\Theta}$. For full derivations, proofs, and explicit formulas, see \cite{kilic_interpretable_2025}.

\subsubsection{Posterior Distribution of Factor Matrices}

Because of the multilinear nature of the CPD, a tensor represented in CPD form can be expressed as a function that is linear with respect to any one of its factor matrices. Thus, data-fitting term $ \bm{U}^{D^T} \bm{w}$ in \eqref{eq5} can be rewritten linearly in terms of the unknown $d$th factor matrix $\mathbf{W}^{(d)}$

\vspace{-0.5cm}
{\fontsize{9}{11}\selectfont
\begin{equation}
    \begin{aligned}
 \bm{U}^{D^T} \bm{w} &= \operatorname{vec}(\mathbf{W}^{(d)})^T \mathbf{G}^{(d)}, \\
\mathbf{G}^{(d)} &= \mathbf{U} \khatri \Big( \circledast_{k \neq d} \mathbf{W}^{(k)T} \mathbf{U}   \Big).
\label{eq13}
\end{aligned}
\end{equation}}

\begin{algorithm}[t!]
\small
\caption{}
\label{alg:btn}
\begin{algorithmic}[1]
    \REQUIRE Input data $\mathbf{x} = \{x_n\}_{n=1}^{N}$, 
             output data $\mathbf{y} = \{y_n\}_{n=1}^{N}$
\STATE \textbf{Initialize:} $R$, $I$, $\mathbf{W}^{(d)}, \mathbf{\Sigma}^{(d)}$, $\forall d \in [1, D]$, 
 $a_0$, $b_0$, $\mathbf{c}_0$, $\mathbf{d}_0$, $\mathbf{g_0}$, $\mathbf{h}_0$ and set $\tau = a_0 / b_0$, $\lambda_r = c_0^r / d_0^r$, $\forall r \in [1, R]$, $\lambda_{I} = g_0^i/h_0^i$, $\forall i \in [1, I]$.
\REPEAT
    \FOR{$d = 1$ to $D$}
        \STATE Update posterior $q_d(\operatorname{vec}(\mathbf{W}^{(d)}))$ using Eq.~\eqref{eq15}
    \ENDFOR
    \STATE Update posterior $q(\bm{\delta})$ using Eq.~\eqref{eq17}
    \STATE Update posterior $q(\bm{\lambda})$ using Eq.~\eqref{eq16}
    \STATE Evaluate variational lower bound 
    \IF{truncation criterion satisfied}
        \STATE Reduce rank $R$ by removing zero-columns in $\mathbf{W}^{(d)}$, $\forall d \in [1,D]$
    \ENDIF
\UNTIL{convergence}
    \STATE Update posterior $q(\tau)$ using Eq.~\eqref{eq19}
\STATE Compute predictive distribution using Eq.~\eqref{eq21}
\end{algorithmic}
\end{algorithm}

where $\mathbf{G}^{(d)} \in \mathbb{R}^{IR\times N}$ is the design matrix. By applying \eqref{eq12}, the posterior mean $\operatorname{vec}(\tilde{\mathbf{W}}^{(d)})$ and covariance $\mathbf{\Sigma}^{(d)}$ of 
{\fontsize{9}{11}\selectfont
\begin{equation*}
q_{\mathbf{W}^{(d)}}(\operatorname{vec}(\mathbf{W}^{(d)})) = 
\mathcal{N}\Big(\operatorname{vec}(\mathbf{W}^{(d)}) \mid \operatorname{vec}(\tilde{\mathbf{W}}^{(d)}), \mathbf{\Sigma}^{(d)}\Big),
\label{eq14}
\end{equation*}
}
are updated by
{\fontsize{9}{11}\selectfont
\begin{equation}
    \begin{aligned}
    \operatorname{vec}(\tilde{\mathbf{W}}^{(d)}) &= \mathbb{E}_q[\tau] \, \mathbf{\Sigma}^{(d)} \, \mathbb{E}_q[\mathbf{G}^{(d)}] \, \mathbf{y}, \\
    \mathbf{\Sigma}^{(d)} &= \Big[ \mathbb{E}_q[\tau] \, \mathbb{E}_q[\mathbf{G}^{(d)} \mathbf{G}^{(d)T}] + 
    \mathbb{E}_q[\bm{\Lambda}] \otimes \mathbb{E}_q[\bm{\Delta}] \Big]^{-1}.
    \end{aligned}
        \label{eq15}
\end{equation}
}

\subsubsection{Posterior distributions of $\bm{\lambda}$ and $\bm{\delta}$}

The posterior of the row precision $\bm{\delta}$ is an independent Gamma distribution over rows $q_{\bm{\delta}}(\bm{\delta}) = \prod_{i=1}^{I} \text{Ga}(\delta_i \mid g_N^i, h_N^i)$ where the posterior parameters are updated by

\vspace{-0.5cm}

{\fontsize{9}{11}\selectfont
\begin{equation}
    \begin{aligned}
    g_N^i = g_0^i + \frac{DR}{2}, \qquad 
    h_N^i = h_0^i + \sum_{d=1}^D \mathbb{E}_q\left[ \mathbf{w}_{i}^{(d)T} \bm{\Lambda} \mathbf{w}_{i}^{(d)} \right].
    \end{aligned}
    \label{eq17}
\end{equation}
}

The posterior expectation of $\bm{\delta}$ can be obtained by $\mathbb{E}_q[\bm{\delta}] = [g_N^1 / h_N^1, ..., g_N^I/h_N^I]^T$, and thus $\mathbb{E}_q[\bm{\Delta}] = \operatorname{diag}(\mathbb{E}_q[\bm{\delta}])$. 

The posterior of $\bm{\lambda}$ is an independent Gamma distribution over columns $q_{\bm{\lambda}}(\bm{\lambda}) = \prod_{r=1}^R \text{Ga}(\lambda_r \mid c_N^r, d_N^r)$ where the posterior parameters are updated by

\vspace{-0.5 cm}

{\fontsize{9}{11}\selectfont
\begin{equation}
    \begin{aligned}
    c_N^r = c_0^r + \frac{DI}{2}, \qquad
    d_N^r = d_0^r + \frac{1}{2} \sum_{d=1}^D \mathbb{E}_q\left[ \mathbf{w}_r^{(d)T} \bm{\Delta} \mathbf{w}_r^{(d)} \right].
    \end{aligned}
    \label{eq16}
\end{equation}
}

\vspace{-0.25cm}

The posterior expectation of $\bm{\lambda}$ can be obtained by $\mathbb{E}_q[\bm{\lambda}] = [c_N^1 / d_N^1, ..., c_N^R/d_N^R]^T$, and thus $\mathbb{E}_q[\bm{\Lambda}_R] = \operatorname{diag}(\mathbb{E}_q[\bm{\lambda}_R])$. 

\subsubsection{Posterior distribution of noise precision $\tau$}

The noise precision $\tau$ is inferred by combining information from the observed data, all factor matrices, and its hyperprior. Its variational posterior is a Gamma distribution $q_{\tau}(\tau) = \text{Ga}(\tau \mid a_N, b_N),$ where the posterior parameters are updated by

\vspace{-0.5cm}
{\fontsize{9}{11}\selectfont
\begin{equation}
\begin{aligned}
        a_N = a_0 + \frac{N}{2}, \qquad 
b_N = b_0 + \frac{1}{2} \mathbb{E}_q\big[\|\mathbf{y} -  \bm{U}^{D^T} \bm{w}\|_F^2\big].
\end{aligned}
\label{eq19}
\end{equation}}

The posterior expectation of $\tau$ is then $\mathbb{E}_q[\tau] = a_N / b_N$. 

\subsubsection{Lower Bound of Model Evidence}

The variational inference framework maximizes the evidence lower bound (ELBO) $\mathcal{L}(q)$, which is guaranteed not to decrease across iterations and can therefore serve as a convergence criterion. It is defined as $\mathcal{L}(q) = \mathbb{E}_{q(\Theta)} \big[\ln p(\mathbf{y}, \Theta)\big] + H(q(\Theta))$, where the first term is the posterior expectation of the joint distribution and the second term is the entropy of the variational posterior distributions.

\subsection{Predictive Distribution}

The predictive distribution for unseen data, given training data, is approximated using the variational posterior as

\vspace{-0.4cm}
{\fontsize{9}{11}\selectfont\begin{equation}
    p(\tilde{y}_i \mid \mathbf{y}) 
\simeq \int \int 
p \big(\tilde{y}_i \mid \{\mathbf{W}^{(d)}\}, \tau^{-1} \big) \, 
q(\{\mathbf{W}^{(d)}\}) \, q(\tau) \, 
d\{\mathbf{W}^{(d)}\} \, d\tau.
\label{eq21}
\end{equation}}

This yields a Student's t-distribution $ \tilde{y}_i \mid \mathbf{y} \sim \mathcal{T}(\tilde{y}_i, \mathcal{S}_i, \nu_y)$, with parameters

\vspace{-0.4cm}
{\fontsize{9}{11}\selectfont
\begin{equation}
\begin{aligned}
    \tilde{y}_i &= \mathop{\circledast}_{d=1}^D \tilde{\mathbf{W}}^{(d)} \mathbf{\varphi}_i^{(d)}, 
\quad \nu_y = 2 a_N, \\
    \mathcal{S}_i &= \left\{ \frac{b_N}{a_N} 
    + \sum_d \mathbf{g}^{(d)}(x_n)^T \mathbf{\Sigma}^{(d)} \mathbf{g}^{(d)}(x_n) \right\}^{-1}.
\end{aligned}
\label{eq23}
\end{equation}
}

The predictive variance can be obtained from the Student's t-distribution as $\operatorname{Var}(y_i) = \frac{\nu_y}{\nu_y - 2} \mathcal{S}_i ^{-1}$. \textbf{Algorithm~1} summarizes the BTN-V presented above.

\section{Numerical Experiments} 

\begin{figure}[t]
  \centering
   \includegraphics[width=0.9\linewidth]{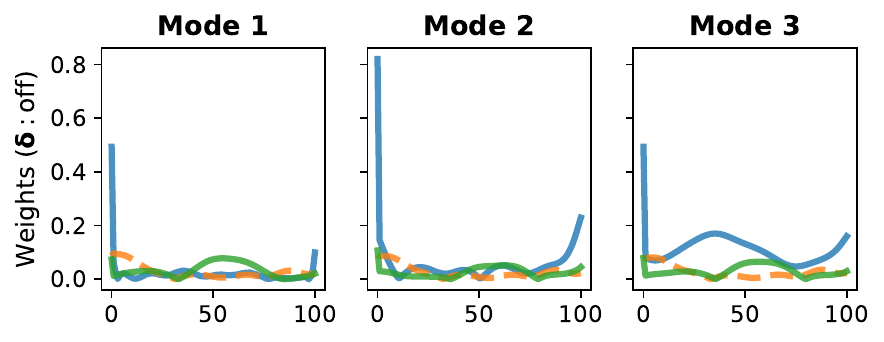}
    \includegraphics[width=0.9\linewidth]{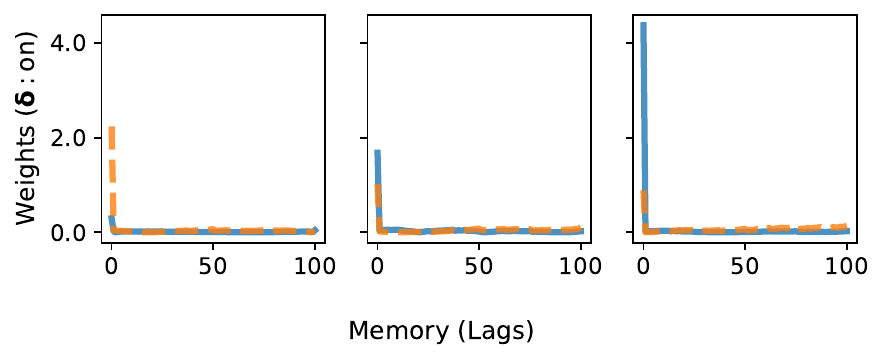}
   \includegraphics[width=0.65\linewidth]{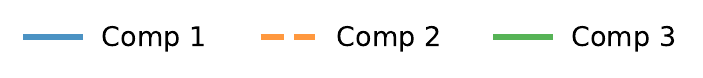}
  \caption{\footnotesize Absolute values of the column weights of the factor matrices without (top) and with (bottom) $\boldsymbol{\delta}$ regularization. The regularization of $\boldsymbol{\delta}$ enforces decay across lags, consistent with the fading memory property of Volterra kernels.}
  \label{fig:delta_reg}
\end{figure}

In this section, we evaluate the performance of BTN-V in terms of predictive accuracy, uncertainty quantification, and computational efficiency, measured by the Root Mean Squared Error (RMSE), Negative Log-Likelihood (NLL), and training runtime, respectively. The NLL is defined as $\mathrm{NLL} = -\tfrac{1}{N} \sum_{n=1}^{N} \log p(y_n \mid \theta)$, and it quantifies how well the predicted probability distribution fits the observed data, penalizing both inaccurate and overconfident predictions. 

In the Volterra model, increasing the memory length ($M$) and polynomial degree ($D$) increases the number of parameters to be learned. Especially when the dataset is short, there are not enough samples to estimate these parameters effectively, which can lead to poor predictive performance. Using the short Cascaded Tanks Benchmark dataset~\citep{Schoukens2016}, we first demonstrate that regularization via $\boldsymbol{\delta}$ and $\boldsymbol{\lambda}$ yields a more parsimonious model with fewer effective parameters and improved predictive accuracy. We then compare the performance of BTN-V with three state-of-the-art methods for Volterra system identification: BMVALS \citep{memmel_bayesian_2023},  RVS \citep{birpoutsoukis_efficient_2018} and SED-MPK \citep{DallaLibera2021}

For all models, the Volterra order is set to \( D = 3 \) and the memory length to \( M = 100 \). For BMVALS, which is a tensor-based Volterra kernel method, we use a rank of \( R = 48 \), as specified in the original paper. For the BTN-V models, the initial CPD rank is set to \( R = 20 \), and the hyperparameters are initialized following the procedure described in \cite{kilic_interpretable_2025}. To ensure numerical stability during identification, all input data in the BTN-V models are normalized to the range \([0, 1]\), and the output data are standardized to have zero mean and unit variance. The normalization on the targets is removed for prediction, and the validation performance of each method is reported. All computations are performed on an Apple MacBook Pro with an Apple M2 Pro chip and 16 GB of RAM running macOS 15.7.1. The Python code enabling the reproduction of all experiments in this section is available at \url{github.com/afrakilic/BTN_Volterra_Sys_ID}.

\subsection{Benchmark Description}

The Cascaded Tanks Benchmark is a nonlinear system with a short dataset (\( N = 1024 \)). The setup includes two water tanks: water is pumped from a reservoir into the upper tank, flows into the lower tank, and then returns to the reservoir. The input signal controls the pump, while the output signal measures the water level in the lower tank. When too much water is pumped from the reservoir, the tanks may overflow. Thus, this dataset exhibits both soft and hard nonlinearities, where the soft nonlinearities arise from the smooth hydraulic flow between the tanks and the hard nonlinearities are caused by tank overflow. In the following, we model the system’s input–output relationship using a Volterra series.

\begin{figure}[t]
  \centering
   \includegraphics[width=0.8\linewidth]{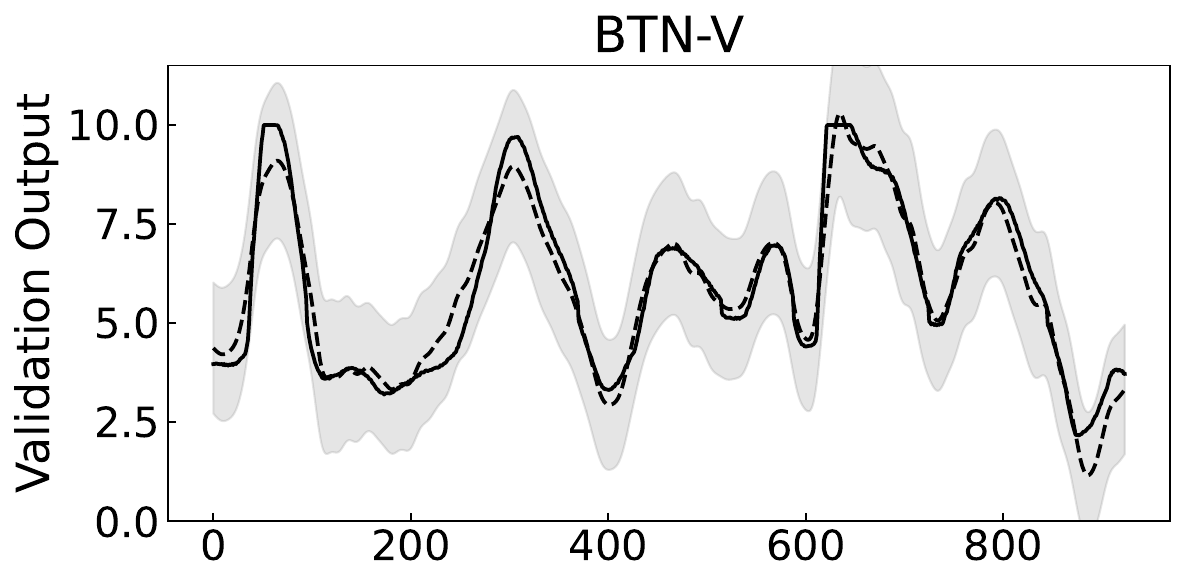}
   \includegraphics[width=0.8\linewidth]{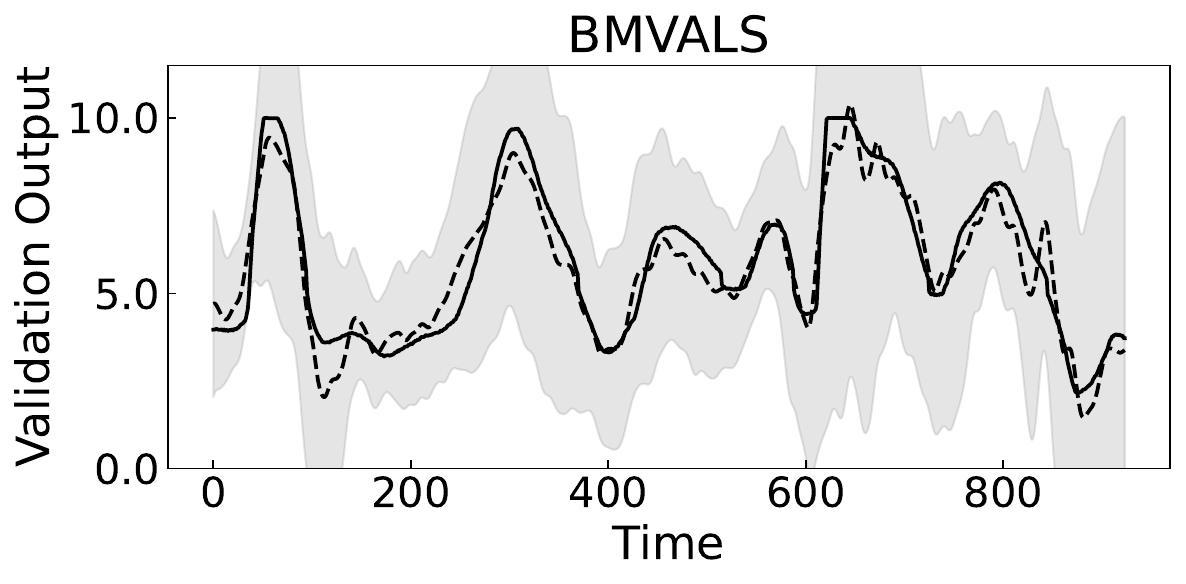}
      \includegraphics[width=0.9\linewidth]{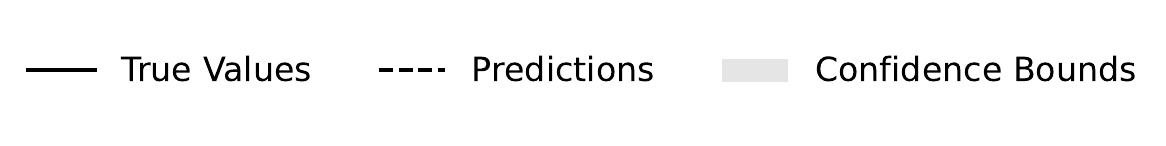}
\caption{\footnotesize Predicted and actual validation outputs for BTN-V and BMVALS. The solid line shows the true observations, the dashed line shows the predictive mean, and the shaded area represents $\pm 3$ standard deviations from the mean.}

  \label{fig:bmvals}
\end{figure}

\subsection{Regularization Through $\bm{\delta}$ and $\bm{\lambda}$}

First to examine the BTN-V's ability to perform automatic rank inference, we analyze whether column-wise regularization of the factor matrices via $\bm{\lambda}$ promotes convergence to low-rank solutions. Since the factor matrices are randomly initialized, we evaluated the model under 10 different random initializations to account for this randomness. With the initial rank fixed at $R=20$, BTN-V yields an average final rank of $2.5 \pm 0.5$, demonstrating convergence to low-rank solutions.

Next, we present the effect of row-wise regularization through $\boldsymbol{\delta}$, which is associated with memory (lags). In the Volterra framework, past observations are expected to have a diminishing influence on the current state. To examine the effect $\boldsymbol{\delta}$ on the factor matrices, we train the models both when the row-wise penalization term $\boldsymbol{\delta}$ is enabled and disabled. Specifically, setting $\boldsymbol{\delta} = \bm{I}$ and keeping it fixed throughout the identification process results in no penalization on the rows; we refer to this setting as $\boldsymbol{\delta}$ : off. Conversely, when $\boldsymbol{\delta}$ is updated during identification, we denote it as $\boldsymbol{\delta}$ : on. We present weights in the resulting factor matrices columnwise in Figure~\ref{fig:delta_reg}. In Figure~\ref{fig:delta_reg}, we plot the factor matrix weights columnwise. For $D = 3$ modes, there are three factor matrices of size $I \times R$, where $I = 101$ for a memory length of $M = 100$. Each column represents a CP component of the Volterra kernel coefficients, and each row corresponds to a specific memory element. The top row of Figure~\ref{fig:delta_reg} shows the column weights without $\boldsymbol{\delta}$ regularization, where no clear decay with increasing lag is observed. This is most evident in the first component (blue line), which even increases at higher lags, while other components exhibit similarly irregular patterns. In contrast, with $\boldsymbol{\delta}$ regularization (bottom row), the estimated rank decreases to 2, and the column weights show a clear decaying trend across lags, as expected for Volterra kernels with fading memory. Unlike RVS and SED-MPK, which rely on fixed exponentially decaying priors, the proposed BTN-V model learns fading memory behavior directly from the data through $\boldsymbol{\delta}$ regularization. As shown in Table 1, applying row-wise regularization $\boldsymbol{\delta}$ also results in lower RMSE and NLL values, indicating improved predictive performance. 

\subsection{Predictive Performance}

Table 1 presents a comparison between the proposed BTN-V and the three state-of-the-art methods on validation data: BMVALS,  RVS and SED-MPK in terms of RMSE, NLL, and computation time. Since the tensor components of  BTN-V and BMVALS are randomly initialized, results are reported as the mean ± standard deviation over 10 runs to account for this randomness.

As mentioned before, RVS and SED-MPK use fixed exponentially decaying priors to model the fading-memory property of Volterra kernels, which improves their predictive accuracy and generally leads to lower RMSE. As shown in Table~1, BTN-V achieves a similar RMSE ($0.51 \pm 0.02$) to SED-MPK ($0.48$) and better than RVS ($0.54$), while being much faster, requiring only $13.68\,\mathrm{s}$ for identification time compared to $120\,\mathrm{s}$ for SED\text{-}MPK and $23{,}400\,\mathrm{s}$ for RVS. Furthermore, we compare BTN-V with BMVALS, a probabilistic tensor-based Volterra kernel method. BTN-V has an average effective rank of $2.5 \pm 0.5$ over 10 runs, while BMVALS uses a fixed rank of $R = 48$, making BTN-V a much simpler model. Despite this, BTN-V achieves a lower RMSE ($0.51 \pm 0.02$ vs.~$0.66$) and yields smoother predictive curves, as shown in Figure~\ref{fig:bmvals}. For uncertainty quantification, BTN-V also reports a lower NLL ($0.77 \pm 0.05$ vs.~$1.23 \pm 0.00$). As illustrated in Figure~\ref{fig:bmvals}, BTN-V provides tighter, more consistent uncertainty bounds, and is about three times faster than BMVALS (13.68~s vs.~38.99~s).

\begin{table}[t] 
\centering 
\vspace{0.2em} 
\begin{tabular}{lccc}
\toprule
\textbf{Method} & \textbf{RMSE} & \textbf{NLL} & \textbf{Time (s)} \\ 
\midrule 
 BTN-V ($\bm{\delta}$: on) & 0.51 $\pm$ 0.02 & \textbf{ 0.77 $\pm$ 0.05} & 13.68 $\pm$ 1.73 \\ 
  BTN-V ($\bm{\delta}$: off) & 0.69 $\pm$ 0.04 & 1.12 $\pm$ 0.10 & \textbf{12.52 $\pm$ 2.12} \\ 
BMVALS & 0.66 $\pm$ 0.00 & 1.23 $\pm$ 0.00 & 38.99 $\pm$ 1.46 \\
RVS* & 0.54 & NA& $\approx$ 23400 \\ 
SED-MPK** & \textbf{0.48} & NA& $\approx$ 120 \\ 
\bottomrule \end{tabular} 
\caption{\footnotesize Performance comparison of BTN-V and existing methods in terms of RMSE, NLL, and computation time. \textbf{Bold} indicates the best result. *RVS and **SED-MPK values are taken from \citep{birpoutsoukis_efficient_2018, DallaLibera2021}.}

\label{tab:Table1} 
\end{table}

\section{Conclusion}

This paper presents a probabilistic method for nonlinear system identification using the TN SISO Volterra series. Using sparsity-inducing hierarchical priors, BTN-V automatically determines the effective tensor rank and fading-memory behavior from data. Experiments on the Cascaded Tanks Benchmark show that BTN-V offers competitive accuracy, improved uncertainty quantification, and reduced computational cost. Future work will focus on extending BTN-V to MIMO systems. In this study, we initialized $\boldsymbol{\Delta} = \boldsymbol{I}$ as an uninformative prior and showed that $\boldsymbol{\Delta}$ learned the fading-memory behavior directly from data. We also plan to explore alternative prior structures, such as decaying priors, and investigate representing the Volterra coefficients using other tensor-network formats, such as tensor-train decomposition, instead of CPD.
 
\bibliography{ifacconf}

@article{chen_estimation_2011,
  author    = {T. Chen and H. Ohlsson and L. Ljung},
  title     = {On the estimation of transfer functions, regularizations and Gaussian processes -- revisited},
  journal   = {IFAC Proceedings Volumes},
  series    = {18th IFAC World Congress},
  volume    = {44},
  number    = {1},
  year      = {2011},
  month     = {Jan.},
  issn      = {1474-6670}
}

@misc{kilic_interpretable_2025,
  author    = {A. Kilic and K. Batselier},
  title     = {Interpretable Bayesian tensor network kernel machines with automatic rank and feature selection},
  publisher = {arXiv},
  note      = {arXiv:2507.11136},
  year      = {2025},
  month     = {Jul.}
}

@article{memmel_bayesian_2023,
  author    = {E. Memmel and C. Menzen and K. Batselier},
  title     = {Bayesian framework for a MIMO Volterra tensor network},
  journal   = {IFAC-PapersOnLine},
  series    = {22nd IFAC World Congress},
  volume    = {56},
  number    = {2},
  pages     = {7294--7299},
  year      = {2023},
  month     = {Jan.},
  issn      = {2405-8963}
}

@article{batselier_tensor_2017,
  author    = {K. Batselier and Z. Chen and N. Wong},
  title     = {Tensor network alternating linear scheme for MIMO Volterra system identification},
  journal   = {Automatica},
  volume    = {84},
  pages     = {26--35},
  year      = {2017},
  month     = {Oct.},
  issn      = {0005-1098}
}

@article{birpoutsoukis_efficient_2018,
  author    = {G. Birpoutsoukis and P. Z. Csurcsia and J. Schoukens},
  title     = {Efficient multidimensional regularization for Volterra series estimation},
  journal   = {Mechanical Systems and Signal Processing},
  volume    = {104},
  pages     = {896--914},
  year      = {2018},
  month     = {May},
  issn      = {0888-3270}
}

@article{pillonetto_new_2010,
  author    = {G. Pillonetto and G. {De Nicolao}},
  title     = {A new kernel-based approach for linear system identification},
  journal   = {Automatica},
  volume    = {46},
  number    = {1},
  pages     = {81--93},
  year      = {2010},
  month     = {Jan.},
  issn      = {0005-1098}
}

@inproceedings{miao_kernels_2019,
  author    = {P. Miao and C. Qi and Y. Jin and K. Song and T. Yu},
  title     = {Kernels pruning for Volterra digital predistortion using sparse Bayesian learning},
  booktitle = {2019 11th International Conference on Wireless Communications and Signal Processing (WCSP)},
  pages     = {1--6},
  year      = {2019},
  month     = {Oct.}
}

@article{favier2012nonlinear,
  author    = {G. Favier and A. Y. Kibangou and T. Bouillot},
  title     = {Non-linear system modeling and identification using Volterra-PARAFAC models},
  journal   = {International Journal of Adaptive Control and Signal Processing},
  volume    = {26},
  number    = {1},
  pages     = {30--53},
  year      = {2012}
}

@book{neal1996bayesian,
  author    = {R. M. Neal},
  title     = {Bayesian learning for neural networks},
  series    = {Lecture Notes in Statistics},
  volume    = {118},
  publisher = {Springer},
  address   = {New York, NY},
  year      = {1996}
}

@article{Zhao_2015_rank_det,
  author    = {Q. Zhao and L. Zhang and A. Cichocki},
  title     = {Bayesian CP factorization of incomplete tensors with automatic rank determination},
  journal   = {IEEE Transactions on Pattern Analysis and Machine Intelligence},
  volume    = {37},
  number    = {9},
  pages     = {1751--1763},
  year      = {2015}
}

@article{kruskal1977three,
  author    = {J. B. Kruskal},
  title     = {Three-way arrays: Rank and uniqueness of trilinear decompositions, with application to arithmetic complexity and statistics},
  journal   = {Linear Algebra and Its Applications},
  volume    = {18},
  number    = {2},
  pages     = {95--138},
  year      = {1977}
}

@article{winn_variational_2005,
  author    = {J. M. Winn and C. M. Bishop},
  title     = {Variational message passing},
  journal   = {Journal of Machine Learning Research},
  volume    = {6},
  pages     = {661--694},
  year      = {2005}
}

@article{DallaLibera2021,
  author    = {A. Dalla Libera and R. Carli and G. Pillonetto},
  title     = {Kernel-based methods for Volterra series identification},
  journal   = {Automatica},
  volume    = {129},
  pages     = {109686},
  year      = {2021}
}

@inproceedings{Schoukens2016,
  author    = {M. Schoukens and P. Mattsson and T. Wigren and J.-P. Noel},
  title     = {Cascaded tanks benchmark combining soft and hard nonlinearities},
  booktitle = {Proceedings of the Workshop on Nonlinear System Identification Benchmarks},
  year      = {2016},
  month     = {Apr.}
}

@techreport{harshman1970foundations,
  author       = {R. A. Harshman},
  title        = {Foundations of the PARAFAC Procedure: Model and Conditions for an {``Explanatory''} Multi-Mode Factor Analysis},
  institution  = {UCLA Working Papers in Phonetics},
  number       = {16},
  year         = {1970}
}
\end{document}